\documentclass[10pt,twocolumn,letterpaper]{article}

\usepackage{iccv}
\usepackage{times}
\usepackage{epsfig}
\usepackage{graphicx}
\usepackage{amsmath}
\usepackage{amssymb}
\usepackage{multirow}
\usepackage{makecell}
\usepackage{multirow}
\usepackage{booktabs}
\usepackage{authblk}
\usepackage{tikz}

\usepackage[pagebackref=true,breaklinks=true,letterpaper=true,colorlinks,bookmarks=false]{hyperref}

\newcommand\copyrighttext{
\centering \footnotesize \copyright 2021 IEEE. Personal use of this material is permitted. Permission from IEEE must be obtained for all other uses, in any current or future media, including reprinting/republishing this material for advertising or promotional purposes, creating new collective works, for resale or redistribution to servers or lists, or reuse of any copyrighted component of this work in other works.}
\newcommand\copyrightnotice{
\begin{tikzpicture}[remember picture,overlay]
\node[anchor=south,yshift=10pt] at (current page.south) {\fbox{\parbox{\dimexpr\textwidth-\fboxsep-\fboxrule\relax}{\copyrighttext}}};
\end{tikzpicture}
}

\iccvfinalcopy % *** Uncomment this line for the final submission

\def\iccvPaperID{14} % *** Enter the ICCV Paper ID here

\ificcvfinal\fi

\begin{document}
%%%%%%%%% TITLE
\title{Deployment of Deep Neural Networks for Object Detection on Edge AI Devices with Runtime Optimization}

\author[1,2,*]{Lukas~St\"acker}
\author[1,3,*]{Juncong~Fei}
\author[1]{Philipp~Heidenreich}
\author[1]{Frank~Bonarens}
\author[2]{\\Jason~Rambach}
\author[2]{Didier~Stricker}
\author[3]{Christoph~Stiller}
\affil[1]{Stellantis, Opel Automobile GmbH, Germany}
\affil[2]{German Research Center for Artificial Intelligence, Germany}
\affil[3]{Institute of Measurement and Control Systems, Karlsruhe Institute of
Technology, Germany}
\affil[*]{equal contribution}
\affil[ ]{\tt \normalsize lukas.staecker@external.stellantis.com}
\affil[ ]{\tt \normalsize juncong.fei@external.stellantis.com}

\maketitle
\ificcvfinal\thispagestyle{empty}\fi

\copyrightnotice

%%%%%%%%% ABSTRACT
\begin{abstract}
Deep neural networks have proven increasingly important for automotive scene understanding with new algorithms offering constant improvements of the detection performance. 
However, there is little emphasis on experiences and needs for deployment in embedded environments. We therefore perform a case study of the deployment of two representative object detection networks on an edge AI platform. In particular, we consider RetinaNet for image-based 2D object detection and PointPillars for LiDAR-based 3D object detection. 
We describe the modifications necessary to convert the algorithms from a PyTorch training environment to the deployment environment taking into account the available tools. 
We evaluate the runtime of the deployed DNN using two different libraries, TensorRT and TorchScript. In our experiments, we observe slight advantages of TensorRT for convolutional layers and TorchScript for fully connected layers. 
We also study the trade-off between runtime and performance, when selecting an optimized setup for deployment, and observe that quantization significantly reduces the runtime while having only little impact on the detection performance.
\end{abstract}

%%%%%%%%% BODY TEXT
\section{Introduction}
Nowadays, we witness a great success of AI-based object detection algorithms with deep neural network (DNN) models. These find applications in automotive scene understanding for advanced driver assistance systems and automated driving. Many object detection algorithms are well studied and their performance in development conditions is known in the literature. However, deployment aspects of these DNN models on edge AI devices, and embedded systems in general, are often not addressed in scientific papers but only in blog posts in the form of partly very helpful or incomplete notes and hints. In this paper, we intent to present our major findings when deploying two representative DNN models on a widely used edge AI device, the NVIDIA Jetson AGX Xavier~\cite{nvidiaJetsonAGX}. 
As representative DNN models, we chose 2D object detection using RetinaNet~\cite{retinanet2017} and 3D object detection using PointPillars~\cite{pointpillars2019}.
Our findings are presented in a concentrated form and include necessary manipulations of the architecture, that take into account the supported functions of TensorRT or TorchScript.

When deploying DNN models, there are several techniques to  optimize the runtime for deployment \cite{gholami2021survey}, including (1)~the design and manipulation of the DNN model architecture in terms of the model size, depth and width, (2)~pruning techniques to remove neurons, groups of neurons, or filters, which have little impact on the output, and (3)~quantization to change the numerical representation of data and network weights \cite{wu2020integer}. 
These techniques generally result in a trade-off between runtime and performance. 
In this paper, we consider techniques (1) and (3). On the one hand, we perform experiments with varying resolution of the input image for RetinaNet~\cite{retinanet2017}, and with varying the number of pillars and points per pillar for PointPillars~\cite{pointpillars2019}. On the other hand, we study the effect of quantization to half-precision floating-point format and to fixed-point arithmetic.

Our contribution is a case study covering the joint treatment of DNN deployment aspects on a widely used edge AI device and an in-depth experimental evaluation of runtime optimization techniques. 
We are not aware of comparable work.
In the following, we provide a description of the used concepts and tools for DNN deployment on the NVIDIA Jetson AGX Xavier, and present the modified architectures of our selected DNNs. 

\section{Concepts and tools}
As it is common in the DNN research community, we develop, train and evaluate our algorithms in a Python environment, where we have chosen the PyTorch library \cite{paszke2019pytorch}. 
However, when deploying the network on an embedded system for real-time inference, we move to a C++ environment, in which other tools are needed. In this section, we give a short overview on the concepts and tools that have proven useful to us.

\begin{figure*}[ht]
\centering
  \includegraphics[width=2.0\columnwidth]{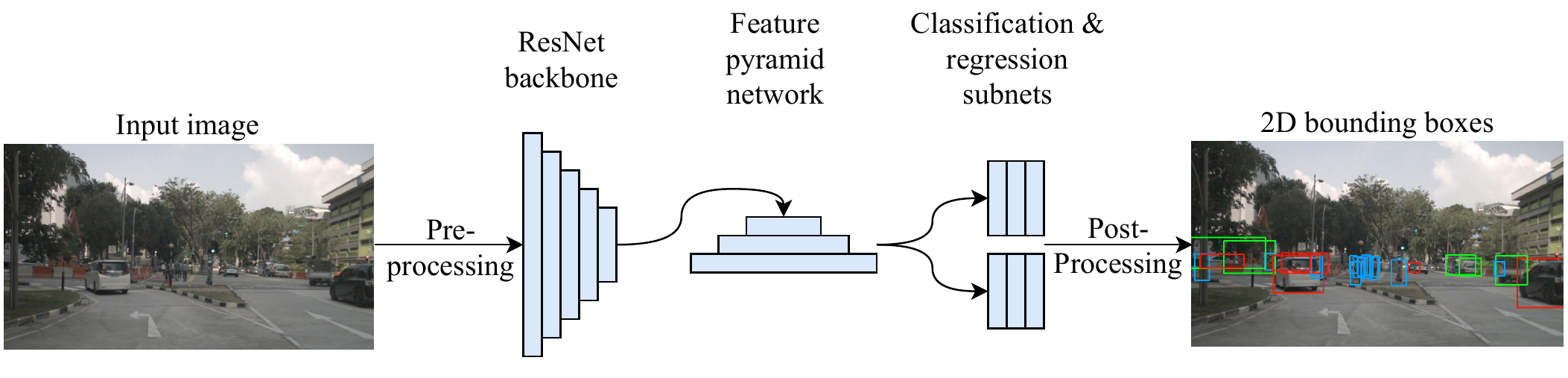}
  \caption{RetinaNet processing pipeline.}
  \label{fig:retinanet_architecture}
\end{figure*}

\subsection{Network conversion}
When deploying a trained DNN on an embedded system in a C++ environment, a conversion becomes necessary. A standardized format to exchange networks within different tools, is the Open Neural Network Exchange (ONNX)~\cite{bai2019onnx} format. It can be created using network tracing, which converts the network into a static computational graph based on exemplary input data and allows for efficient hardware acceleration with the NVIDIA TensorRT~\cite{vanholder2016tensorrt} SDK for runtime optimized inference. However, tracing does not allow data-dependent control-flow operations, so that the original network architecture has to be divided into network graphs and logical operations. These logical operations do not contain network weights and have to be realized with C++ functions. 
PyTorch offers an alternative built-in framework TorchScript, with the options tracing and scripting. For reasons of comparability and a limited usability of scripting, we focus on tracing with TorchScript as well.

\subsection{Quantization}
Quantization has the goal to efficiently represent numerical values by a finite number of bits. 
Widely used formats include single- and half-precision floating-point formats \cite{IeeeFloatNorm}, referred to as Float32 and Float16, respectively, and fixed-point arithmetic with 8 bit integers, referred to as Int8.
A reduction of the used quantization format can be beneficial in a deployment setup with limited computational, memory or energy resources. 
Moving from the default Float32 to Float16 is a straightforward and often considered option. When further moving to Int8 quantization, several tuning and calibration steps have to be taken into account.
A recent survey of quantization techniques for DNN inference is given in \cite{gholami2021survey}. In \cite{wu2020integer}, a corresponding practical workflow for Int8 quantization is recommended.

For the sake of simplicity, in this paper, we focus on post-training quantization and do not consider quantization-aware training. In particular, for Int8 quantization, we consider the TensorRT MinMax and entropy calibrator functions \cite{tensorrt8bit}. The MinMax calibration measures the maximum absolute activation of each layer and provides an equidistant and symmetric mapping. Likewise, the entropy calibration determines a mapping which minimizes the information loss by saturating the activations above a certain threshold. Since TorchScript currently only supports Int8 quantization for CPU usage, we only consider the corresponding TensorRT variant.

\subsection{Further tools}
We use the Robot operating system (ROS)~\cite{quigley2009ros} as our real-time environment in C++. ROS contains helpful tools for sensor data streaming, communication of different nodes and visualization of sensor data and detection results. With the tools nuscenes2bag~\cite{nuscenes2bag} and kitti2bag~\cite{kitti2bag}, we can convert data from automotive datasets into a ROS compatible format that allows us to simulate a real driving scenario while having access to labeled ground truth.
For the pre- and post-processing of the data in C++, efficient implementations from libraries like OpenCV~\cite{opencv_library} for images and OpenPCDet~\cite{openpcdet2020} for pointclouds can be used. For many operations, there are also CUDA-based~\cite{cuda_library} implementations for hardware acceleration in these stages.
%https://github.com/opencv/opencv/wiki/CiteOpenCV

\section{Deployment architectures}
For our case study, we choose to work with two DNN architectures: RetinaNet~\cite{retinanet2017} and PointPillars~\cite{pointpillars2019}. Note that we select them as representative and well-known algorithms, that provide a reasonable trade-off between runtime and performance, and are therefore suitable for automotive applications. Further, note that our selection also covers diversity in terms of the object detection task and and the sensor modality. 

\subsection{RetinaNet}
RetinaNet~\cite{retinanet2017} is an image-based 2D object detection algorithm. It is one of the pioneering one-stage object detectors, which surpassed the preceding two-stage networks in terms of runtime while offering similar detection performance.
When deploying RetinaNet, we divide the pipeline shown in Figure \ref{fig:retinanet_architecture} into the following processing steps:

\textbf{Pre-processing}. 
In the pre-processing stage, the image is resized to a lower resolution of choice and normalized based on the mean and standard deviation of the RGB images in the ImageNet dataset \cite{Krizhevsky.2012}, on which the backbone network is pre-trained. Both of these steps can efficiently be done using the OpenCV CUDA library.

\textbf{Inference}. 
The architecture of RetinaNet~\cite{retinanet2017} is made out of a ResNet~\cite{He.2015} backbone network, a subsequent Feature Pyramid Network (FPN) \cite{Lin.2017b} as well as regression and classification heads. It uses the concept of anchor boxes as pre-defined regions in the image, eliminating the need for a Region Proposal Network.
All of these steps can be realized with a single neural network graph, that can be created by tracing the computations of an exemplary network input. Assuming that all pre-processed images have the same resolution, the anchor boxes can be included in the graph as a constant tensor. The graph can then be efficiently deployed using either TensorRT or TorchScript.

\textbf{Post-processing}. 
In the post-processing stage, the neural network classification and regression outputs are filtered by a detection threshold and decoded to generate the 2D bounding boxes. Finally, non-maximum suppression (NMS) is applied to avoid multiple boxes per object. Again, this can be efficiently computed using the OpenCV library. Note that these steps can not be included in the network graph, as they require logical operations that depend on the network input but would be treated as constant by the tracer.

\subsection{PointPillars}

\begin{figure*}[hbt]
\centering
  \includegraphics[width=2.0\columnwidth]{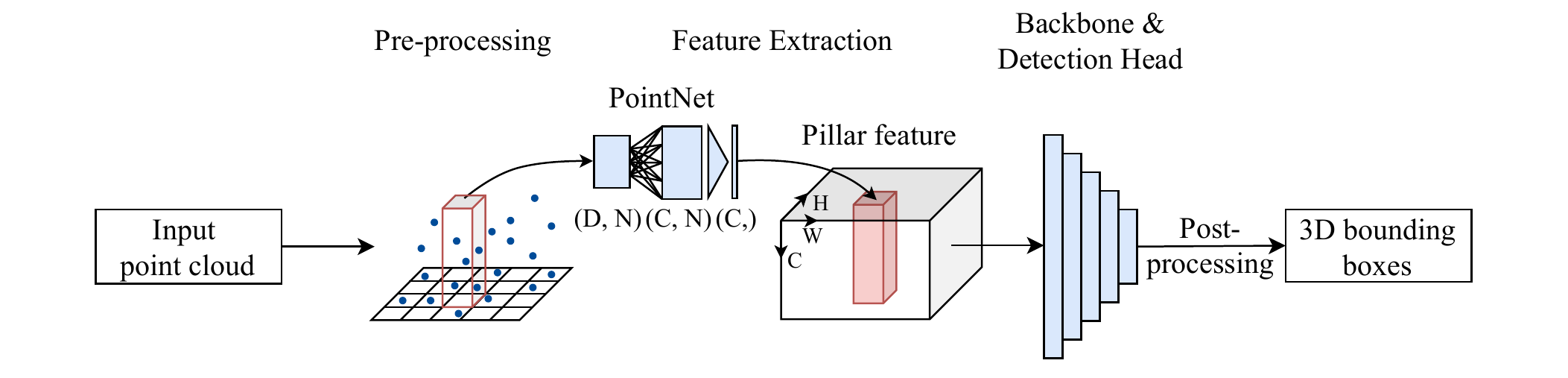}
  \caption{PointPillars processing pipeline.}
  \label{fig:pointpillars_architecture}
\end{figure*}

PointPillars~\cite{pointpillars2019} is a LiDAR-based 3D object detection network that achieves state-of-the-art performance on public benchmarks with real-time inference speed.
As depicted in Figure \ref{fig:pointpillars_architecture}, PointPillars mainly consists of the following four blocks:

\textbf{Pre-processing}. 
In this stage, the input point cloud is first discretized into a set of pillars in the $xy$ plane.
Then, each point in a pillar is augmented with its offsets from the arithmetic mean of all points in this pillar and its offsets from the pillar center, resulting in a $D$-dimensional point encoding. 
For each pillar, we random sample $N$ points if it has more than $N$ points or apply zero-padding to fix the size of the input tensor.
Similarly, the number of non-empty pillars per point cloud is kept to $P$ using the aforementioned policy.
Consequently, the augmented input feature is of size $(P, N, D)$.
The pre-processing block is deployed efficiently with parallel processing on GPU via the CUDA library.

\textbf{Feature Extraction}. 
This block aims to extract high-level features from the input feature obtained in pre-processing, and further form them in a 2D top-view representation.
To this end, each point in pillars is first consumed by a simplified PointNet~\cite{pointnet}, outputing a tensor of size $(P, N, C)$.
Then, a max operation along the $N$ axis is applied to generate pillar-wise feature, resulting in a $(P, C)$ sized tensor.
Finally, these pillar-wise features are scattered back to the pillar locations on the $xy$ plane to create a top-view representation.
In this work, we name the combination of the simplified PointNet and the max operation as Pillar Feature Network~(PFN), and deploy it using TensorRT or TorchScript network graphs.
The scatter operation is realized by C++ functions using the CUDA library.

\textbf{Backbone and Detection Head}. 
The backbone used in PointPillars~\cite{pointpillars2019} is common 2D CNN and consists of a top-down module that gradually captures higher semantic information and a second module that performs upsampling and feature concatenation.
The final feature map is processed by the detection head for anchor classification, box offsets regression and direction regression.
During deployment, the backbone and detection head are jointly realized by TensorRT or TorchScript network graphs.

\textbf{Post-processing}.
To get the final 3D bounding boxes, the predicted box offsets and directions are decoded together with the anchor information. Then, NMS is applied to select the best predictions out of overlapping boxes.
We realize these two logical operations with CUDA-based C++ functions to leverage parallel processing on GPU.

\section{Experiments}
We train and evaluate our algorithms in a Python environment using the PyTorch library. As training and validation data, we use the  nuScenes dataset \cite{caesar2020nuscenes} for RetinaNet~\cite{retinanet2017} and the KITTI dataset \cite{geiger2013vision} for PointPillars~\cite{pointpillars2019}. We deploy the networks in a C++ environment with ROS for data streaming and visualization. We measure the algorithm runtime as the average runtime across the validation data.
We experiment with different tools for hardware acceleration and quantization on the target platform, TensorRT and TorchScript. For the inference with TensorRT, we first export the trained PyTorch model to ONNX and parse it to an optimized TensorRT runtime engine in a C++ environment on the target system. TensorRT allows us to select the desired quantization when building the engine, with the options of Float32, Float16, and Int8. The latter requires a use-case specific calibration, which can be done with the MinMax or entropy technique. With TorchScript, we need to select the quantization before exporting the PyTorch model. Currently, it is only possible to run Float32 and Float16 calculations with hardware acceleration.
In addition, we experiment with varying the input image resolution for RetinaNet and the number of pillars and points per pillar for PointPillars~\cite{pointpillars2019}.
Finally, we study the impact of the available power supply on the runtime of our entire detection pipeline.

\subsection{RetinaNet}
\textbf{Experimental Setup}.
For our experiments, we choose to work with RetinaNet-18~\cite{retinanet2017}, which is based on a small Resnet-18~\cite{He.2015} backbone to account for our real-time requirements. We train and evaluate the network with the nuScenes \cite{caesar2020nuscenes} {\it train} and {\it val} dataset, respectively. 
To adapt the original dataset for the task of 2D object detection, we project the original 3D bounding boxes onto the image plane and map the 27 classes to a reduced set of {\it Car}, {\it Pedestrian}, {\it Truck}, {\it Motorcycle} and {\it Bicycle}. We list both the mean average precision (mAP), with IoU threshold of 0.5, and the weighted mAP, which takes the number of objects per class as weights and is more robust to changes in rare classes.

\textbf{Framework Analysis}.
In the first experiment, we study the runtime of RetinaNet~\cite{retinanet2017} when deploying it with TensorRT and TorchScript, using the available quantization techniques. We consider two different inference batch sizes, corresponding to a single image and a full surround view of six images, as present in autonomous vehicle prototypes or the used nuScenes dataset. 
Table \ref{table:retinanet-quantization} shows the results of this experiment. When running the model with the default Float32 values, we can observe that the inference with TensorRT is substantially faster than with TorchScript, which underlines its optimization capabilities for convolutional layers. 
When using Float16 precision, we observe a significant reduction of the inference time across both tools, while TensorRT still allows for faster inference. With Int8 quantization in TensorRT, we again significantly reduce the runtime, resulting in over four times faster inference relative to Float32.
Some studies have shown that the runtime reduction can have even greater effects when working with larger batch sizes, offering more room for optimization \cite{tensorrt8bit}. In our experiment, we can confirm this trend when running inference on images from all six cameras in the nuScenes dataset in a single batch, observing the reduction factors slightly rising for Int8 quantization with TensorRT.

\textbf{Runtime-Performance Analysis}.
Of course, we are not only interested in the runtime reduction but rather in the trade-off between runtime and detection performance. In the next experiment, we therefore study the trade-off achievable with quantization versus changing the resolution of the input image.
The results of this experiment are shown in Table \ref{table:retinanet-resolution}. We compare low, mid and high resolution, which are chosen to have about twice as many pixels as the preceding resolution. Note that the number of anchor boxes increases proportional to the number of pixels. All models have been trained with input data down-scaled to the desired resolution. The mid resolution is the same that we used in the first experiment and we chose to work with TensorRT for this experiment due to the lower runtime and availability of Int8 quantization.

The runtime reduction factor offered by quantization is similar in all dimensions with almost three times faster inference with Float16 and about four times faster inference with Int8. Interestingly, the impact of Float16 quantization on the performance is minimal, both in terms of mAP and weighted mAP. This also holds for Int8 quantization, where the performance is slightly reduced for the MinMax calibrator.  
We assume that this is due to the 8-bit RGB images as input data, eliminating the need for higher-precision calculations. 
The impact of the image resolution is high on both runtime and detection performance, where the runtime decreases proportional to the number of pixels.
We therefore argue that the mid resolution model with Int8 quantization offers the best trade-off between runtime and performance.

\textbf{Power Supply Analysis}.
In a final experiment for RetinaNet~\cite{retinanet2017}, we study the impact of the available power on the runtime of the model. We select the mid resolution model and Int8 quantization from the previous experiment due to its good runtime-performance trade-off. All experiments so far have been conducted with the MAXN power mode of the NVIDIA Jetson AGX, which corresponds to roughly 50\,W. However, in a practical deployment setup, the power resources are usually limited. 
As shown in Table~\ref{table:retinanet-powermode}, reducing the power mode has a high impact on the resulting runtime of the entire detection pipeline. We therefore list detailed results of pre-processing, inference and post-processing run-times. Depending on the use-case, a trade-off between power supply and runtime has to be made.

\textbf{Summary}.
In conclusion, the experiments show that TensorRT achieves lower inference times for RetinaNet~\cite{retinanet2017} that TorchScript. While quantization should always be considered for deployment, lowering the resolution can help to further reduce the runtime, but at the cost of a decreased detection performance. It is therefore advisable to prefer a mid to high resolution and Int8 quantization over a low resolution and Float32 precision. The runtime is also heavily influenced by the available power supply.
A video showing qualitative results on a nuScenes sequence obtained by this detection pipeline is available at:\\
\href{https://youtu.be/b6\_HPLOUW9I}{https://youtu.be/b6\_HPLOUW9I}.

\begin{table}[hbt]
\caption{Runtime evaluation of RetinaNet with TensorRT and TorchScript using various quantization techniques and batch sizes.\vspace{2mm}}
\label{table:retinanet-quantization}
\centering
    \begin{tabular}{c|c|r|r}
    Batch size & Quant. & TensorRT & TorchScript \\
    \hline
\multirow{3}{*}{1} & Float32                      & 104\,ms  & 210\,ms     \\
& Float16                      & 38\,ms   & 67\,ms            \\
& Int8                         & 25\,ms  & - \\
\hline
\multirow{3}{*}{6} & Float32                      & 619\,ms  & 1041\,ms     \\
& Float16                      & 201\,ms   & 271\,ms            \\
& Int8                         & 136\,ms  & - 
\end{tabular}
\end{table}

\begin{table}[hbt]
\caption{Performance and runtime evaluation of RetinaNet using different image resolutions and quantization techniques on the nuScenes val set.\vspace{2mm}}
\label{table:retinanet-resolution}
\centering
\resizebox{\columnwidth}{!}{%
    \begin{tabular}{c|c|r|c|c}
    Resolution & Quant. & Runtime & mAP & weighted mAP \\
    \hline
\multirow{3}{*}{\makecell{Low \\ 416$\times$736}} & Float32 & 60\,ms & 0.298 & 0.391\\
    & Float16 & 22\,ms & 0.299 & 0.391\\
    & Int8, Entropy & \multirow{2}{*}{16\,ms} & 0.298 & 0.391\\
    & Int8, MinMax & & 0.288 & 0.381\\
    \hline
\multirow{3}{*}{\makecell{Mid \\ 576$\times$1024}} & Float32 & 104\,ms & 0.361 & 0.455\\
    & Float16 & 38\,ms & 0.356 & 0.454\\
    & Int8, Entropy & \multirow{2}{*}{25\,ms} & 0.355 & 0.455\\
    & Int8, MinMax & & 0.355 & 0.454\\
    \hline
\multirow{3}{*}{\makecell{High \\ 832$\times$1472}} & Float32 & 224\,ms & 0.395 & 0.488\\
    & Float16 & 74\,ms & 0.395 & 0.487\\
    & Int8, Entropy & \multirow{2}{*}{50\,ms} & 0.393 & 0.485\\
    & Int8, MinMax & & 0.388 & 0.483 
\end{tabular}%
}
\end{table}

\begin{table}[hbt]
\caption{Detailed runtime of each block in the optimal RetinaNet detection pipeline using different power modes.\vspace{2mm}}
\label{table:retinanet-powermode}
\centering
    \begin{tabular}{l|c|c|c|r}
     Block & MAXN & 30W & 15W & 10W \\
    \hline
Pre-process                    & 3\,ms & 5\,ms & 6\,ms & 7\,ms \\
Inference                     & 24\,ms & 35\,ms & 45\,ms & 94\,ms \\
Post-process                    & 4\,ms & 4\,ms & 5\,ms & 6\,ms  \\
    \hline
Total                     & 31\,ms & 44\,ms & 56\,ms & 107\,ms \\
\end{tabular}
\end{table}

\subsection{PointPillars}
\textbf{Experimental Setup}. 
We train PointPillars~\cite{pointpillars2019} and then evaluate the performance of the deployed network on the KITTI~\cite{kitti} 3D object detection dataset, which contains 7481 image and point cloud pairs with available 3D bounding box annotations for three categories: {\it Car}, {\it Pedestrian} and {\it Cyclist}.
For the experimental evaluation, we follow~\cite{fei2020semanticvoxels} and split the samples into a {\it train} and {\it val} set with 3712 and 3769 samples, respectively. 
The performance is evaluated using Average Precision (AP) in moderate level for the 3D object detection task with Intersection over Union~(IoU) thresholds of 0.7 for {\it Car} and 0.5 for {\it Pedestrian} and {\it Cyclist}.
To construct and train the model, we use the same KITTI-PointPillars configuration as in the \mbox{OpenPCDet} codebase~\cite{openpcdet2020}.

For the deployment on the target platform, we implement the pipeline based on Autoware~\cite{kato2018autoware} by developing additional features. The runtime measurement is conducted on the target platform, if not specified otherwise, with the MAXN power mode using the sequence 0095 with 236 frames in the KITTI raw data~\cite{geiger2013vision}.
We average the runtime starting from the tenth frame to consider the warm-up of GPU on the target platform and thus eliminate occasionality.

\textbf{Runtime Analysis}. 
The runtime results for both Pillar Feature Network (PFN) and Backbone \& Detection Head (named as 2D CNN for short) when deployed with TensorRT and TorchScript using various quantization techniques are presented in Table~\ref{table:pointpillars_quantization}.
%TRT vs TS
When comparing the runtime of the model deployed using TensorRT and TorchScript, PFN and 2D CNN show opposite trends. For the 2D CNN, TensorRT achieves significantly better runtime optimization, which is consistent with our observations when deploying RetinaNet. In contrast, for the PFN, TorchScript provides an improved runtime optimization.
One possible explanation could be slight advantages of TensorRT for convolutional layers and TorchScript for fully connected layers.
% FP32 vs FP16 vs Int8
When comparing the quantization techniques, Float16 offers a significant speed improvement for the PFN and 2D CNN across both tools, whereas Int8 only provides additional minor improvements for the 2D CNN. 
Note that we do not consider to use Int8 quantization for the PFN, since the input are 3D coordinates, with approximate range $\pm10^2\,\mathrm{m}$ and accuracy $10^{-2}\,\mathrm{m}$, so that Int8 quantization would result in a significant loss of information.

\textbf{Performance Analysis}. 
From the runtime analysis, we can conclude that the combination of PFN with TorchScript and 2D CNN with TensorRT achieves the lowest runtime.
We therefore consider this setup and further study the object detection performance of the model deployed using different quantization techniques.
The results of the performance evaluation for three classes as well as the runtime for the entire pipeline are shown in Table~\ref{tab:pointpillars_performance}.
% fp32, fp16
When going down from Float32 to Float16 precision, we observe no significant performance change in terms of AP for all class categories for both PFN and 2D CNN. 
% Int8
When using Int8 quantization for the 2D CNN, the performance drops considerably for both calibration methods, where the reduction is even more severe for the entropy calibrator. We assume that the used pillar feature map representation and subsequent activations require a larger arithmetic range to preserve essential geometric information.
When further considering the runtime of the complete PointPillars~\cite{pointpillars2019} detection pipeline, it is clear that deploying PFN and 2D CNN using Float16 precision with TorchScript and TensorRT, respectively, offers the best trade-off between detection performance and runtime. We name it the optimal variant in the following study.

\textbf{Further Study on PFN}. 
From the previous analysis we observe that the optimal variant takes in total 41\,ms runtime, where the single PFN costs 19\,ms, being the bottleneck in the deployment.
We thus conduct this experiment to investigate the potential of further reducing the runtime of PFN by modifying the network input.
This is achieved by adjusting two parameters of PFN, namely the number of non-empty pillars $P$, and the number of points per pillar $N$.
While we use $P\!=\!16000$ and $N\!=\!32$ in the previous experiments, we additionally choose $P\!=\!12000$ and $N\!\in\!\{24, 16\}$ and measure the runtime of PFN and the overall performance. 
As shown in Table~\ref{tab:pfn_optimization}, keeping the number of points while lowering the number of pillars from 16000 to 12000 does not introduce notable change on the performance metrics, while it improves the runtime considerably from 19\,ms to 14\,ms for the case with 32 points per pillar.
Additional runtime boost can be achieved by decreasing the number of points per pillar without remarkable performance loss.
In this study, we figure out that the setup with 12000 pillars and 24 points per pillar is the most appropriate for deploying PFN with a runtime of 9\,ms.

\textbf{Summary}. 
From our extensive studies, we summarize that deploying PointPillars~\cite{pointpillars2019} using separate PFN and 2D CNN network graphs on Float16 precision with Torchscript and TensorRT, respectively, is the optimal solution for the deployment on the target platform.
We further observe that the input parameters of PFN have a significant impact on the runtime, where we find $P\!=\!12000$ and $N\!=\!24$ as most suitable in our study.
We finally report the detailed runtime of the optimal PointPillars~\cite{pointpillars2019} pipeline using various power modes on the target platform in Table~\ref{tab:pointpillars_runtime}.
A video showing qualitative results on the KITTI sequence obtained by this detection pipeline is available at:\\
\href{https://youtu.be/RL8-3toRTeg}{https://youtu.be/RL8-3toRTeg}.

\begin{table}[hbt]
\centering
\caption{Runtime of PFN and 2D CNN deployed with TensorRT and TorchScript using various quantization techniques.\vspace{2mm}}
\label{table:pointpillars_quantization}
\begin{tabular}{c|c|c|c}
Network & Quant. & TensorRT & TorchScript \\
\hline
\multirow{2}{*}{PFN} & Float32 & 30\,ms  & 21\,ms     \\
& Float16                      & 26\,ms   & 19\,ms            \\
\hline
\multirow{3}{*}{2D CNN} & Float32 & 46\,ms  & 82\,ms     \\
& Float16                      & 16\,ms   & 31\,ms            \\
& Int8                         & 14\,ms  & - 
\end{tabular}
\end{table}

\begin{table}[hbt]
\centering
\caption{Performance evaluation of PFN and 2D CNN combinations using different quantization techniques on the KITTI val set.\vspace{2mm}}
\label{tab:pointpillars_performance}
\resizebox{\columnwidth}{!}{%
\begin{tabular}{c|c|c|c|c|c}
\multicolumn{1}{c|}{\multirow{2}{*}{\begin{tabular}[c]{@{}c@{}}PFN \\quant.\end{tabular}}} & \multicolumn{1}{c|}{\multirow{2}{*}{\begin{tabular}[c]{@{}c@{}}2D CNN \\quant.\end{tabular}}} & \multicolumn{3}{c|}{$\mathrm{AP}_{\mathrm{3D}}$}  & \multirow{2}{*}{\begin{tabular}[c]{@{}c@{}}Pipeline\\runtime\end{tabular}}  \\ 
\cline{3-5}
\multicolumn{1}{l|}{} & \multicolumn{1}{l|}{} & Car & Ped. & Cyc. & \multicolumn{1}{l}{} \\ 
\hline
Float32 & Float32 & 78.40 & 51.41 & 62.81 & 72\,ms \\ 
\hline
Float32 & Float16 & 78.30 & 51.31 & 62.89 & 43\,ms \\ 
\hline
\multirow{2}{*}{Float32} & Int8, minmax & 71.04 & 48.00 & 55.94 & \multirow{2}{*}{40\,ms} \\ 
\cline{2-5}
 & \multicolumn{1}{l|}{Int8, entropy} & 68.91 & 22.03 & 29.66 & \\          
\hline
Float16 & Float32 & 78.40 & 51.46 & 62.92 & 71\,ms \\ 
\hline
Float16 & Float16 & 78.31 & 51.39 & 62.97 & 41\,ms \\
\hline
\multirow{2}{*}{Float16} & Int8, minmax & 70.99 & 47.65 & 56.42 & \multirow{2}{*}{38\,ms} \\ 
\cline{2-5}
 & Int8, entropy & 69.37 & 21.86 & 29.53 & \\
\end{tabular}%
}
\end{table}

\begin{table}[hbt]
\centering
\caption{Performance and runtime evaluation of PFN with different parameters.\vspace{2mm}}
\label{tab:pfn_optimization}
\begin{tabular}{c|c|c|c|c|r}
\multirow{2}{*}{\begin{tabular}[c]{@{}c@{}}Pillar \\num.\end{tabular}} & \multirow{2}{*}{\begin{tabular}[c]{@{}c@{}}Point \\num.\end{tabular}} & \multicolumn{3}{c|}{$\mathrm{AP}_{\mathrm{3D}}$} & \multirow{2}{*}{\begin{tabular}[c]{@{}c@{}}PFN \\runtime\end{tabular}}  \\ 
\cline{3-5}
   &   & Car   & Ped.  & Cyc.  &  \\ 
\hline
16000 & 32 & 78.31 & 51.39 & 62.97 & 19\,ms \\ 
\hline
16000 & 24 & 78.28 & 50.68 & 62.20 & 12\,ms \\ 
\hline
16000 & 16 & 78.14 & 50.81 & 61.33 & 10\,ms \\ 
\hline
12000 & 32 & 78.29 & 51.37 & 62.98 & 14\,ms \\ 
\hline
12000 & 24 & 78.25 & 51.92 & 62.19 & 9\,ms \\ 
\hline
12000 & 16 & 78.17 & 49.02 & 61.38 & 8\,ms 
\end{tabular}
\end{table}

\begin{table}[hbt]
\centering
\caption{Detailed runtime of each block in the optimal PointPillars detection pipeline using different power modes.\vspace{2mm}}
\label{tab:pointpillars_runtime}
\begin{tabular}{l|r|r|r|r}
\multicolumn{1}{c|}{Block} & \multicolumn{1}{c|}{MAXN} & \multicolumn{1}{c|}{30W} & \multicolumn{1}{c|}{15W} & \multicolumn{1}{c}{10W}  \\ 
\hline
Pre-process & 4\,ms & 6\,ms & 8\,ms & 14\,ms \\
PFN & 9\,ms & 14\,ms & 21\,ms & 52\,ms \\
Scatter & 1\,ms & 2\,ms & 2\,ms & 4\,ms \\
2D CNN & 16\,ms & 25\,ms & 32\,ms & 71\,ms \\
Post-process & 1\,ms & 1ms & 2\,ms & 2\,ms \\
\hline
Total & 31\,ms & 48\,ms & 65\,ms & 143\,ms                 
\end{tabular}
\end{table}

\section{Conclusion}
In this paper, we have presented our major insights when deploying two representative algorithms on the NVIDIA Jetson AGX Xavier for automotive scene understanding: RetinaNet~\cite{retinanet2017} for image-based 2D object detection and PointPillars~\cite{pointpillars2019} for LiDAR-based 3D object detection. We have discussed the necessary modifications and tools that we found most helpful. We studied the runtime of TensorRT and TorchScript and found that TensorRT should be preferred for RetinaNet~\cite{retinanet2017} and the convolutional part of PointPillars~\cite{pointpillars2019}, whereas TorchScript should be preferred for the fully connected part of PointPillars~\cite{pointpillars2019}. 
The runtime of both algorithms can be further reduced by utilizing quantization, which is available up to Int8 with TensorRT and up to Float16 with TorchScript. While the impact on the detection performance for RetinaNet~\cite{retinanet2017} is low even with Int8, PointPillars~\cite{pointpillars2019} should only be quantized to Float16, which is possibly due to the difference in input data, with 8-bit RGB images and 3D coordinates of LiDAR point clouds, respectively. 
We also studied the influence of some design parameters of the algorithms and found that a good runtime-performance trade-off can be achieved with an input resolution of 576$\times$1024 for RetinaNet~\cite{retinanet2017}, as well as 12000 pillars and 24 points per pillar for PointPillars~\cite{pointpillars2019}. The available power supply in the embedded environment also has a significant impact on the runtime, which additionally has to be considered when choosing a setup for deployment.
In future work, we plan to extend our findings on fusion methods between image and LiDAR or radar point cloud data. Additionally, we consider to address pruning techniques for further runtime optimization.

\section*{Acknowledgment}
The research leading to these results is partly funded by the German Federal Ministry for Economic Affairs and Energy within the project “Methoden und Maßnahmen zur Absicherung von KI basierten Wahrnehmungsfunktionen für das automatisierte Fahren (KI-Absicherung)". The authors would like to thank the consortium for the successful cooperation.

{\small
\bibliographystyle{ieee_fullname}
\bibliography{refs}
}

\end{document}